\documentclass{dexgem-report}
\usepackage{amsmath,amssymb}
\usepackage{multirow}
\usepackage{threeparttable}
\usepackage{placeins}

\title{Dexterous Robot Manipulation from Human Demonstrations via Contact-Anchored Retargeting and Residual Policy Learning}
\reportdate{September 2026}

\author{Zihao Yang\affil{1}\textsuperscript{*} \and Chengyuan Liu\affil{1}\textsuperscript{*} \and Yu Zhou\affil{1,2} \and Runze Lv\affil{1,2} \and Tianyu Cui\affil{1,3} \and Sheng Yi\affil{1,4} \and Haohua Zhu\affil{1,4} \and Irvine Lu\affil{1}\textsuperscript{\ensuremath{\dagger}} \and JieQ Sun\affil{1}\textsuperscript{\ensuremath{\dagger}}}
\affiliation{1}{DexGEM Lab}
\affiliation{2}{Shanghai Jiao Tong University}
\affiliation{3}{Tongji University}
\affiliation{4}{DexRobot Co.\ Ltd.}
\authornote{\textsuperscript{*}Equal contribution.\quad \textsuperscript{\ensuremath{\dagger}}Corresponding authors: Irvine Lu (irvine@dex-gem.ai), JieQ Sun (jieq@dex-gem.ai).}
\codelink{https://github.com/DexGEM-Lab/real2sim2real}

\begin{document}
\maketitle

\begin{abstract}
Learning dexterous manipulation from demonstrations is
bottlenecked by data: the contact forces that determine whether a
grasp succeeds are absent from every scalable source of human
demonstrations. This paper builds on two observations. First, what
survives the change from a human hand to a robot hand is the
contact structure of a demonstration---which finger regions touch
which object locations, and in what order---rather than its joint
motion. Second, physical consistency need not be engineered per
task: a single residual reinforcement learning (RL) policy, trained
once across diverse demonstrations, can repair kinematic recordings
into physically consistent, contact-annotated trajectories, and the
same residual formulation restores dynamic feasibility after
retargeting. These observations yield a three-stage pipeline that
converts human motion-capture recordings into dexterous robot
policies with no real-robot training data: physics refinement with
a simulated MANO hand recovers contacts and forces,
contact-anchored retargeting transfers the demonstrated contact
structure through an objective independent of hand morphology, and
residual policy learning adapts the result to robot actuation. The
pipeline reconstructs 25{,}454 single-hand trajectories (success
7.3\%$\to$59.3\%) and 25 dual-hand tasks (16.0\%$\to$62.4\%) with
one shared policy per setting, transfers one human dataset to four
morphologically distinct robot hands (+62.4~pp), and executes four
contact-rich bimanual tasks on physical hardware with zero
real-robot training data.
\end{abstract}

\section{Introduction}

\begin{figure}[t!]
\centering
\includegraphics[width=\linewidth]{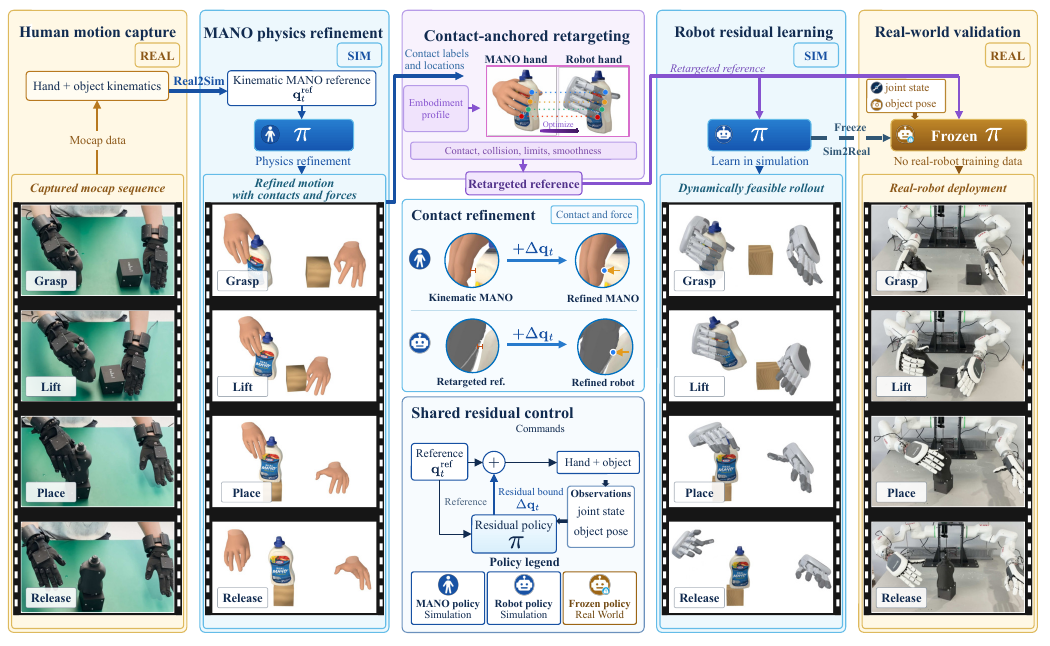}
\caption{Pipeline overview. \textbf{Human motion capture} records
 hand and object kinematics at 120\,Hz in a room-scale mocap system.
 \textbf{MANO physics refinement} re-executes the kinematic reference
 with a residual-RL policy in simulation, recovering per-frame
 contacts and forces (contact refinement insets).
\textbf{Contact-anchored retargeting} adjusts the robot hand so that the
 designated contact regions on its fingers reach the same object-surface
 locations as in the human demonstration, transferring the contact
 pattern rather than copying joint angles.
 \textbf{Robot residual learning} restores dynamic feasibility
 via a second residual policy trained in simulation. Both sides
 share the residual control formulation. \textbf{Real-world validation}
 freezes the policy and transfers it to physical hardware with zero
 real-robot training data, using joint-state and object-pose feedback.}
\label{fig:intro}
\end{figure}

Dexterous manipulation, the ability to grasp, reposition, and
reorient objects with a multi-fingered hand, depends on precisely
coordinated finger--object contact. Learning from demonstrations is
the prevailing route to such skills, avoiding the per-task reward
engineering of reinforcement
learning~\cite{andrychowicz2020dexterous, chen2022reorientation},
but it shifts the bottleneck to data: contact-rich demonstrations
are precisely what is hard to collect at scale.

Robot teleoperation~\cite{handa2020dexpilot, qin2023anyteleop,
arunachalam2023dime, ding2024bunnyvisionpro,
cheng2024opentelevision, fu2024mobilealoha} is the de facto
standard for demonstration data and has produced the field's
largest manipulation datasets~\cite{oneill2024openx}. It works where the operator's command space
matches the robot's action space, as for arms with parallel-jaw
grippers; consistent with this, the 22~embodiments of Open
X-Embodiment are overwhelmingly
gripper-based~\cite{oneill2024openx}. For multi-fingered hands,
three problems compound: human hand control is organized in
low-dimensional synergies rather than independent joints, so
operators cannot deliberately steer 20+ coupled degrees of freedom
through a retargeting layer they cannot
perceive~\cite{santello1998synergies, li2026survey}; sensing and
retargeting latency degrades operator control
quality~\cite{pattiwar2025latency}; and
operators receive no haptic feedback for contacts their own fingers
occlude, where removing force feedback alone drops contact-rich
teleoperation success from 78\% to 48\%~\cite{gao2025glovity}.

Video offers the cheapest human data: multi-view reconstruction
converts human clips into robot demonstrations~\cite{qin2022dexmv},
and egocentric video scales collection by orders of
magnitude~\cite{kareer2024egomimic, zheng2026egoscale}. Video
alone, however, recovers neither precise hand--object pose nor
contact force; EgoScale's action labels themselves come from
instrumented gloves rather than pixels~\cite{zheng2026egoscale}.
Motion capture occupies the precision end, recording
millimeter-accurate hand and object kinematics without a robot in
the loop~\cite{dexcap2024}. Yet the recording remains purely
kinematic: the contact forces that distinguish stable grasps from
infeasible poses~\cite{brahmbhatt2019contactgrasp} are unrecorded,
and the human hand's morphology prevents direct replay on a
robot~\cite{qin2022onehand}.

Converting such demonstrations into deployable robot skills
therefore requires closing two gaps. The \emph{embodiment gap} is
the question of what transfers. The instinctive answer, mapping
joint angles or fingertip positions, treats hands with different
kinematics as scaled copies of one another and fails precisely when
morphology differs~\cite{qin2022onehand}; contact-aware
alternatives produce only static grasps~\cite{wang2023dexgraspnet}
or pay a heavy per-trajectory sampling cost~\cite{pan2025spider}.
What actually survives the change of embodiment is the contact
structure: which finger regions touch which object locations, and
in what order. The \emph{physics gap} is the question of what makes
the motion feasible. A kinematic recording contains no forces, so
reconstruction in simulation is unavoidable; but existing
real-to-sim-to-real (real2sim2real) pipelines treat reconstruction
as per-task engineering, with a separate policy, reward curriculum,
and training run for every new
skill~\cite{qin2022dexmv, zhao2025dexmachina, xu2026demobot}. The
reconstruction machinery, not the data, then becomes the scaling
bottleneck.

This paper presents a three-stage pipeline that closes both gaps
with no teleoperated data at any stage (Fig.~\ref{fig:intro}).
\emph{Physics refinement} re-executes each recording in simulation
with a simulated MANO hand~\cite{romero2017mano}, letting the
physics engine supply the missing contact forces; the key design is
a single residual-RL policy~\cite{johannink2019residual,
silver2018residual} trained jointly across all demonstrations,
which repairs kinematics rather than learning tasks, and therefore
needs no per-task training and can synthesize additional physically
consistent variants of each demonstration.
\emph{Contact-anchored retargeting} transfers the demonstrated
contact structure to a target robot hand: the optimization asks
that designated finger surface regions reach the recorded contact
locations on the object, and its objective depends only on surface
geometry, not on the hand's kinematics. \emph{Residual policy
learning} restores dynamic feasibility on the robot by learning
small corrections to the retargeted reference, reusing the same
residual formulation as physics refinement, so that the resulting
policy transfers directly to hardware~\cite{peng2018simtoreal}.

Our contributions are:
\begin{itemize}
\item \textbf{Contact structure as the transferable content of a
  demonstration.} We identify per-frame contact labels and
  object-surface contact locations, rather than joint
  trajectories, as the embodiment-invariant content of a
  demonstration, and derive a retargeting objective mathematically
  independent of hand morphology; a single human dataset transfers
  to four morphologically distinct robot hands with identical
  configuration (Sec.~\ref{sec:exp_cross}).
\item \textbf{Physics consistency as an amortized, learned
  operator.} We show that a single residual-RL policy, trained once
  across all tasks, converts raw kinematic recordings into
  physically consistent, contact-annotated data, removing per-task
  reward engineering from the real2sim2real loop and generalizing
  to unseen objects without fine-tuning (Sec.~\ref{sec:exp_r2s}).
\item \textbf{End-to-end validation at scale.} We validate on
  25{,}454 single-hand and 250 dual-hand trajectory
  reconstructions, downstream policy gains from recovered contact
  forces (+11.8~pp), and real-robot execution of four
  contact-rich bimanual dexterous tasks with a single residual
  policy shared across tasks, zero real-robot training data, and no
  per-task reward design or retraining (Sec.~\ref{sec:exp_real}).
\end{itemize}

\section{Related Work}
\label{sec:related}

\subsection{Learning Dexterous Manipulation from Human Data}

Teleoperation systems map human motion onto dexterous hand-arm
platforms~\cite{handa2020dexpilot, qin2023anyteleop,
arunachalam2023dime, ding2024bunnyvisionpro, cheng2024opentelevision}
and bimanual gripper systems~\cite{fu2024mobilealoha}, but require a
skilled operator on the physical hardware; adding force or tactile
feedback helps contact-rich operation~\cite{gao2025glovity}, but
adds hardware and system complexity.
Human-side capture avoids the robot during collection:
DexCap~\cite{dexcap2024} replays mocap data through kinematic
retargeting, DexMV~\cite{qin2022dexmv} converts multi-view video
into robot demonstrations, UMI~\cite{chi2024umi} transfers
in-the-wild demonstrations to grippers, and
HumanPlus~\cite{fu2024humanplus} shadows human motion onto
humanoids. Egocentric video scales collection
further~\cite{kareer2024egomimic}, with
EgoScale~\cite{zheng2026egoscale} pretraining dexterous policies
on over 20{,}000 hours of egocentric video.
Overall, these pipelines either require per-task training or
operate in kinematic space, discarding the contact forces that
govern manipulation success.

\subsection{Hand Retargeting}

Classical retargeting optimizes geometric correspondences, such as
fingertip distance vectors~\cite{handa2020dexpilot} or joint-angle
and fingertip-position objectives~\cite{qin2023anyteleop}. These
correspondences assume compatible kinematic topology and degrade
when finger count, proportions, or joint ranges
differ~\cite{qin2022onehand}. Contact- and force-closure-based
synthesis~\cite{wang2023dexgraspnet} adapts across hands but
produces static grasps; SPIDER~\cite{pan2025spider} converts
kinematic references into dynamically feasible trajectories through
physics-based sampling at high per-trajectory cost;
UniDexGrasp~\cite{xu2023unidexgrasp} learns goal-conditioned
grasping but stops at acquisition and lifting. Our contact-anchored
retargeting extends contact-based transfer from grasp acquisition
to temporally coherent manipulation sequences.

\subsection{Real-to-Sim-to-Real Pipelines}

Sim-to-real reinforcement
learning~\cite{andrychowicz2020dexterous, peng2018simtoreal}
avoids real-robot training, and real2sim2real pipelines reconstruct
demonstrations in simulation, refine policies there, and transfer
back. Two problems recur. The first is per-task training:
DexMV~\cite{qin2022dexmv} trains a separate RL policy per task,
DexMachina~\cite{zhao2025dexmachina} designs curriculum rewards
per skill, and DexMimicGen~\cite{jiang2025dexmimicgen} still
trains per-task imitation policies on generated data. Most closely
related, DemoBot~\cite{xu2026demobot} applies residual RL to
motion priors from human video but requires per-task temporal
segmentation, an event-driven reward curriculum, and a
success-gated reset strategy. In contrast, our physics-refinement
policy is trained once across all demonstrations. The second is
limited task complexity: existing systems demonstrate in-hand
rotation~\cite{chen2022reorientation}, gripper
pick-and-place~\cite{fu2024mobilealoha}, or
grasping~\cite{xu2023unidexgrasp}, and diffusion-policy
methods~\cite{chi2023diffusion} operate at the gripper level.
Real2sim2real pipelines have yet to address contact-rich bimanual
dexterous manipulation, where two multi-fingered hands coordinate
on a shared task; we validate on four such tasks with zero per-task
fine-tuning.

\section{Method}

The pipeline converts a human motion-capture recording into a
robot manipulation policy in three stages
(Fig.~\ref{fig:intro}). \emph{Physics refinement}
(Sec.~\ref{sec:physics}) reproduces captured hand and object motion in
physics simulation, recovering per-frame contact points and forces.
\emph{Contact-anchored retargeting} (Sec.~\ref{sec:retarget})
transfers the demonstrated contact structure from the human hand onto a
target robot hand. \emph{Residual policy learning}
(Sec.~\ref{sec:residual}) restores dynamic feasibility on the robot
embodiment. Sec.~\ref{sec:deploy} describes hardware
deployment.

Throughout this paper, we use the MANO hand
model~\cite{romero2017mano} as the human-side representation and the
DexHand~021 \cite{yuan2026dexhand}---a tendon-driven five-finger
hand with 12 active controls coupled to 20 physical finger joints---as
the experimental robot hand. The method itself is not specific to this
robot: changing the target hand requires only an updated embodiment
profile (Sec.~\ref{sec:retarget}) and retraining the robot-side
residual policy.

\subsection{Motion Capture and Data Processing}
\label{sec:capture}

\begin{figure}[!t]
\centering
\includegraphics[width=\linewidth]{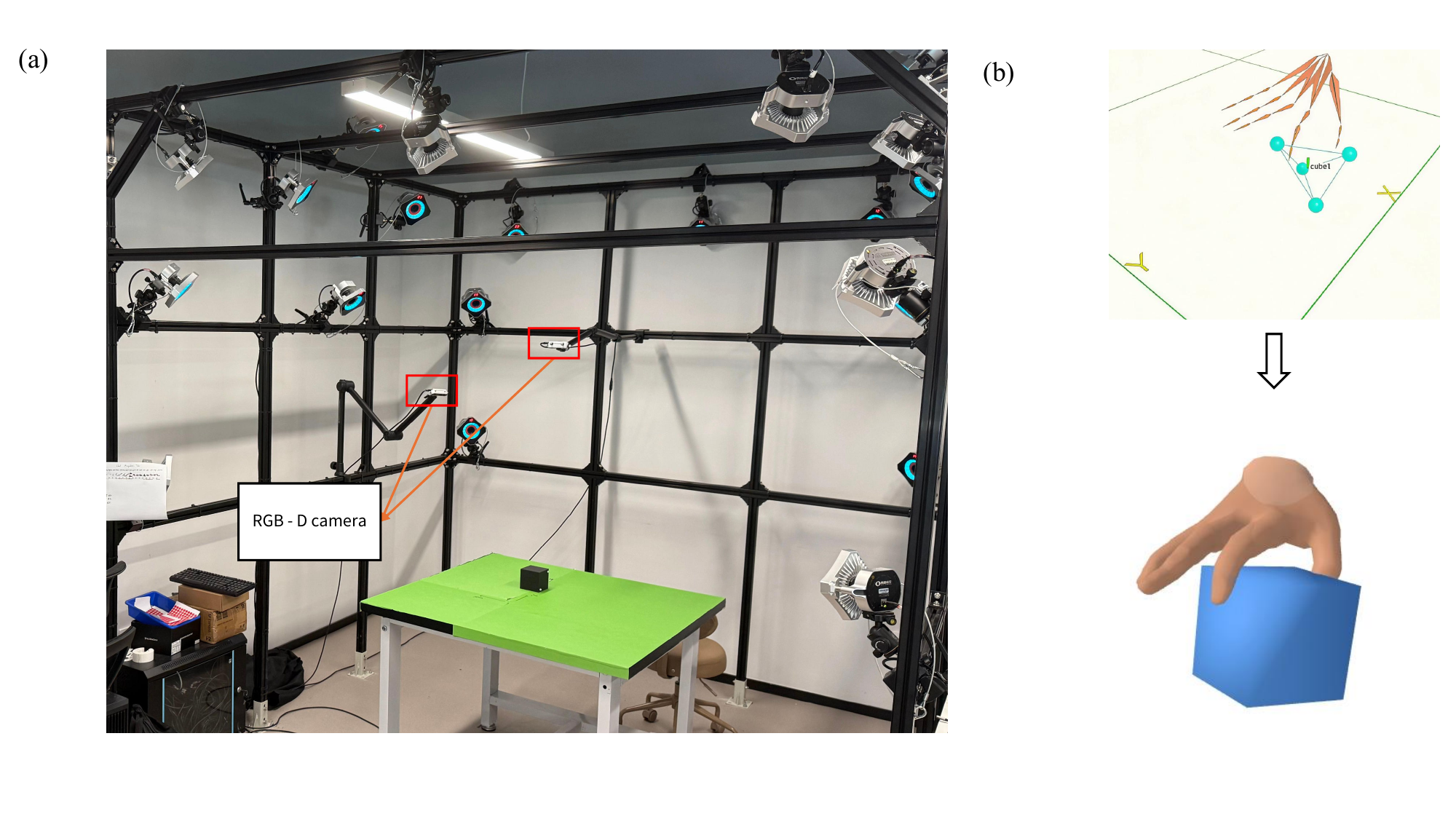}
\caption{Data capture and processing.
  \textbf{(a)}~Room-scale optical motion-capture system with
  22~infrared cameras and synchronized RGB-D sensors.
  \textbf{(b)}~Processing pipeline: raw mocap markers are
  pre-processed into the MANO coordinate frame, fitted to the MANO
  hand model~\cite{romero2017mano}, and passed through physics
  simulation to obtain contact force annotations.}
\label{fig:capture}
\end{figure}

The input to our pipeline is a set of motion-capture recordings of
human manipulation (Fig.~\ref{fig:capture}). Demonstrations are
captured in a room-scale system equipped with 22~infrared cameras
for millimeter-precision optical tracking and two synchronized
RGB-D cameras. The operator's hand carries
14~reflective markers at anatomical landmarks; each object carries
markers on 3D-printed mounts whose positions correspond exactly
to the URDF model used in simulation, eliminating coordinate-frame
misalignment. The system records hand pose (6-DoF wrist and finger
joints) and object trajectories at 120\,Hz.

Raw marker trajectories are transformed into the MANO coordinate
system via a calibrated world-to-MANO rigid transform. Per-subject
hand shape parameters ($\beta \in \mathbb{R}^{10}$) are estimated
from calibration sequences using HaMeR~\cite{pavlakos2024hamer}
applied to synchronized RGB frames, and frame-wise wrist SE(3) and
finger pose are optimized to minimize marker-to-surface distances
under anatomical constraints. The resulting MANO trajectories
(Fig.~\ref{fig:capture}b), together with the tracked object
trajectories, constitute the kinematic input to the physics
refinement stage described next.

\subsection{Physics Refinement}
\label{sec:physics}

Physics refinement recovers the contact forces missing from raw
kinematic recordings by re-executing each demonstration in
Isaac Gym~\cite{makoviychuk2021isaacgym} with a simulated actuated
MANO hand model, letting the simulator's contact model produce
physically consistent forces.
A residual-RL policy~\cite{johannink2019residual,
silver2018residual} outputs corrections to the captured reference:
\begin{equation}
\Delta\mathbf{q}_t =
\gamma\,\Delta\mathbf{q}_{t-1}
+ \mathbf{s}\odot\mathbf{a}_t,
\qquad
\mathbf{q}^{\mathrm{cmd}}_t
= \mathbf{q}^{\mathrm{ref}}_t
+ \Delta\mathbf{q}_t,
\label{eq:residual}
\end{equation}
where $\mathbf{q}^{\mathrm{ref}}_t$ is the reference pose,
$\mathbf{a}_t$ is the policy action, $\mathbf{s}$ is a per-dimension
scale, and $\gamma$ is a decay coefficient.
The accumulated correction $\Delta\mathbf{q}_t$ is clipped per
dimension, constraining the commanded pose to a bounded envelope
around the reference.
The policy is trained to maximize a reward combining
object-trajectory tracking (penalizing deviation of the simulated
object from its recorded motion) with contact-achievement terms that
encourage designated fingers to reach the object surface. The MANO
hand model is a parametric mesh; for physics simulation we convert
it into an articulated rigid-body model with
22~actuated finger joints and a 6-DoF wrist driven by PD controllers.

\textbf{One policy per setting.} A single policy is trained
jointly over all trajectories within each setting
(single-hand or dual-hand), conditioned on task identifier,
object geometry, and current and future reference object motion.
Because the policy repairs kinematic references rather than planning
autonomously, these observations are available at both training and
inference. This design makes physics refinement a scalable
data-repair operator: any new demonstration is processed without
per-trajectory retraining.

\textbf{Synthesis.} The trained policy can be rolled out under varied
conditions---perturbed initial object poses, randomized friction and
mass, and random external force perturbations---to generate additional
physically consistent variants of each demonstration. Each variant
preserves the task's contact structure while exploring neighboring
dynamically feasible trajectories.

Each physics-refined trajectory stores the MANO hand motion and
object motion together with per-frame \emph{contact labels} (which
finger regions are in contact), \emph{object-surface contact
locations} (where on the object surface each contact occurs), and
\emph{reconstructed force vectors}. The contact labels and locations
are the primary inputs to retargeting.

\subsection{Contact-Anchored Retargeting}
\label{sec:retarget}

For each frame of a physics-refined trajectory, we optimize the
robot hand's configuration
$\mathbf{u}_t \in \mathbb{R}^{D}$---packing wrist SE(3) and the
hand's active finger controls---so that designated fingertip surface
points approach the recorded MANO contact locations on the object.
Frames are segmented by their \emph{contact pattern} (the set of
finger regions simultaneously in contact), and different objectives
apply to multi-contact versus sparse-contact frames.

\textbf{Multi-contact frames} minimize:
\begin{equation}
\begin{split}
E_{\mathrm{ct}} =\;& w_{\mathrm{tgt}} E_{\mathrm{tgt}}
+ w_{\mathrm{pen}} E_{\mathrm{pen}}
+ w_{\mathrm{h2o}} E_{\mathrm{h2o}}
+ w_{\mathrm{spen}} E_{\mathrm{spen}}
+ w_{\mathrm{j}} E_{\mathrm{joint}} \\
&+ w_{\mathrm{g}} E_{\mathrm{ground}}
+ w_{\mathrm{b}} E_{\mathrm{sm}}^{\mathrm{base}}
+ w_{\mathrm{c}} E_{\mathrm{sm}}^{\mathrm{ctrl}}
+ w_{\mathrm{r}} E_{\mathrm{sm}}^{\mathrm{rot}}.
\end{split}
\label{eq:contact}
\end{equation}
The contact-anchor term $E_{\mathrm{tgt}}$ is the primary objective.
Each contact label from the physics-refined trajectory names an
anatomical finger region (e.g., ``index fingertip'' or ``thumb
pad''). On the robot hand, this region maps to a set of mesh surface
points. For each contact, we select the surface point whose direction
from the link center best matches the direction of the corresponding
MANO contact, with distance as tiebreaker. The optimization then
adjusts $\mathbf{u}_t$ so this point approaches the recorded
object-surface position, penalizing distance beyond a 1\,cm tolerance
$\delta$:
\begin{equation}
E_{\mathrm{tgt}} = \sum_{i \in \mathcal{C}_t}
\big[\max\big(\|c_i(\mathbf{u}_t)-p_i\|-\delta,\,0\big)\big]^2.
\label{eq:target}
\end{equation}
The remaining terms penalize object/self-penetration
($E_{\mathrm{pen}}$, $E_{\mathrm{spen}}$), excessive hand--object
distance ($E_{\mathrm{h2o}}$), joint-limit violations
($E_{\mathrm{joint}}$), ground penetration ($E_{\mathrm{ground}}$),
and temporal jitter in wrist and finger controls.

\begin{figure}[t]
\centering
\includegraphics[width=\linewidth]{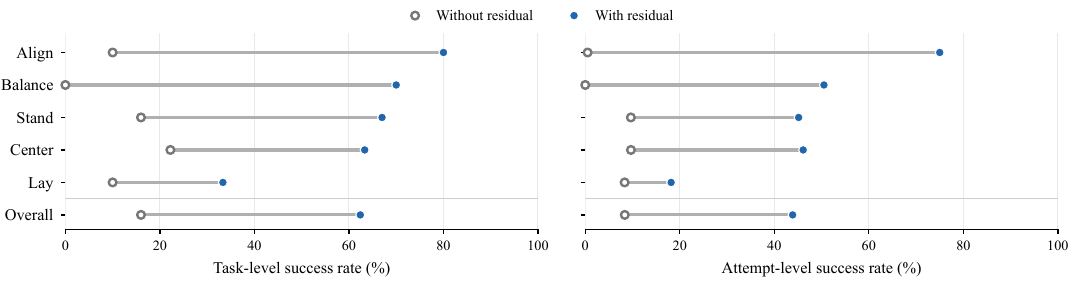}
\caption{Dual-hand evaluation grouped by task type
  (25~tasks, 250~trajectories, $\pm$10\% object-size perturbation).
  Each row aggregates all tasks of a given type; Overall is the
  aggregate across all types. Gray: without residual correction;
  blue: with residual correction. Left: task-level success rate;
  right: attempt-level success rate.}
\label{fig:bimanual}
\end{figure}

\textbf{Sparse-contact frames} (zero or one contact) optimize finger
controls only with feasibility terms, keeping the wrist frozen.

\textbf{Initialization.} Each frame is initialized from the
corresponding MANO configuration. The 22~MANO finger-joint values are
mapped to the robot hand's active finger controls through an
approximate kinematic correspondence, clipped to the robot's joint
limits. Wrist orientation is taken from the MANO wrist relative to
the object; wrist translation is initialized as the mean offset
between robot and MANO fingertip positions.

\textbf{Morphology generality.} The optimization operates on
forward-kinematic surface points, embodiment geometry, joint limits,
and an active control vector. Changing the target hand changes the
optimization dimension and geometric queries but not the mathematical
loss. An \emph{embodiment profile}---specifying the kinematic model,
control-to-joint coupling, joint limits, and candidate surface points
per contact region---is the only hand-specific input.

\subsection{Residual Policy Learning}
\label{sec:residual}

The retargeted trajectory places the robot's fingers at the correct
contact locations in the correct order, but it is a kinematic
solution: it does not account for actuator torque limits, tendon
coupling, or object dynamics under the robot's contact geometry.
A second residual policy (Eq.~\eqref{eq:residual}) learns small
corrections to wrist translation and finger controls; wrist
orientation follows the reference unchanged. On the DexHand~021,
the policy's 12~active finger controls are routed through the hand's
deterministic tendon coupling to drive 20~physical finger joints.

After the final demonstrated contact, the finger dimensions of
$\gamma$ are set to one (no decay), so accumulated finger corrections
persist and established grasps are maintained; wrist-translation
dimensions retain their original decay schedule.

As with physics refinement, a single policy per setting is trained
across all retargeted trajectories in
Isaac Gym~\cite{makoviychuk2021isaacgym} with
PPO~\cite{schulman2017ppo}, conditioned on task
identifier, object geometry, and reference motion. Synthesis rollouts
under domain randomization~\cite{peng2018simtoreal} produce
additional dynamically feasible robot trajectories from each
retargeted reference.

\subsection{Hardware Deployment}
\label{sec:deploy}
At each control cycle, the policy receives the robot's current joint
positions and object pose from hardware sensors, formatted in the
same observation schema used during training. It computes a residual
correction, adds it to the current frame of the retargeted reference,
and publishes joint commands. Policy weights are frozen.

Before each episode, a static geometric calibration aligns the
hardware coordinate frame with the simulation frame. No
dynamics identification, policy fine-tuning, or online adaptation is
performed.


\section{Experiments}

We organize the experiments around three questions, one per
contribution.
(1)~Does a single amortized policy close the physics gap at scale,
and does it generalize beyond its training distribution?
(Sec.~\ref{sec:exp_r2s})
(2)~Does contact-anchored transfer close the embodiment gap better
than geometric retargeting, in isolation and across morphologies?
(Sec.~\ref{sec:exp_retarget})
(3)~Do the resulting data improve downstream policy learning, and
does the complete pipeline run on real hardware with zero
real-robot training data? (Sec.~\ref{sec:exp_downstream})
All simulation success evaluations share a single criterion
(Sec.~\ref{sec:exp_r2s}); object subscripts are numbered by order
of appearance, independently per table or figure.

\subsection{Physics Consistency at Scale}
\label{sec:exp_r2s}

\textbf{Setup.}
One shared policy is trained per setting with no per-task tuning.
The single-hand dataset contains 25{,}454 mocap trajectories
(14~objects, 85~object--gesture combinations, with objects and
gestures as defined in DexCanvas~\cite{xu2025dexcanvas}); the
dual-hand dataset contains 250~trajectories (25~tasks, 12~objects). During evaluation
we perturb each object--gesture pair by up to 10\% of the object
size and attempt every trajectory 50~times. A rollout succeeds if
no object deviates more than 3\,cm from its mocap reference
throughout execution, and for orientation-constrained tasks the
final-frame error must not exceed $30^{\circ}$.

\textbf{Results.}
Table~\ref{tab:single} reports per-object success rates.
Replay without residual correction averages only 7.26\%, whereas
residual correction raises the average to 59.31\%: every object
improves, and on the four objects where replay never succeeds
(bowl, cuboid$_1$, cylinder$_4$, sphere$_1$), residual correction
reaches 41.5\%--70.8\%.
Results on the dual-hand dataset are consistent
(Fig.~\ref{fig:bimanual}): residual correction raises task-level
success from 16.0\% to 62.4\% and attempt-level from 8.4\% to
43.9\%, with no regression on any task type.
These success rates also reflect data synthesis quality---the
trained policy serves as a data augmentation engine, randomly
perturbing initial object and hand poses within a bounded range
and filtering synthesized trajectories by the same success criteria.

\begin{table}[t]
\centering
\caption{Single-hand success rates with and without residual
  correction (14~objects, 85~object--gesture combinations).
  The average is the unweighted mean over the 14~objects.}
\label{tab:single}
\vspace{2pt}
\footnotesize
\begin{tabular}{@{}lccc@{}}
\toprule
Object & Replay & With residual & Pairs \\
\midrule
Large clamp   & 23.99\% & 62.80\% & 5 \\
Cylinder$_1$  & 16.49\% & 68.20\% & 8 \\
Cube$_1$      & 15.44\% & 77.20\% & 7 \\
Mayonnaise    & 10.70\% & 56.40\% & 5 \\
Banana        &  9.86\% & 59.40\% & 7 \\
Cube$_2$      &  9.12\% & 81.10\% & 8 \\
Cylinder$_2$  &  8.88\% & 66.40\% & 7 \\
Power drill   &  5.85\% & 33.50\% & 6 \\
Cylinder$_3$  &  0.66\% & 35.50\% & 9 \\
iPhone        &  0.61\% & 48.80\% & 8 \\
Bowl          &  0.00\% & 41.50\% & 3 \\
Cuboid$_1$    &  0.00\% & 60.00\% & 5 \\
Cylinder$_4$  &  0.00\% & 68.80\% & 3 \\
Sphere$_1$    &  0.00\% & 70.80\% & 4 \\
\midrule
\textbf{Average} & \textbf{7.26\%} & \textbf{59.31\%} & \textbf{85} \\
\bottomrule
\end{tabular}
\end{table}

\textbf{Unseen object generalization.}
Without any fine-tuning, the same shared policy is evaluated on
six YCB objects~\cite{calli2015ycb} absent from training, as shown
in Table~\ref{tab:ycb}. The pooled rate is 72.4\%, with per-object
rates spanning 46.5\%--88.7\%, confirming generalization to novel
object instances. Such zero-shot transfer distinguishes a learned
repair operator from a per-dataset artifact.

\begin{table}[t]
\centering
\caption{Generalization to unseen YCB objects (zero fine-tuning).}
\label{tab:ycb}
\vspace{2pt}
\footnotesize
\setlength{\tabcolsep}{10pt}
\begin{tabular}{@{}lcc@{}}
\toprule
YCB Object & Trials & Rate \\
\midrule
Tomato soup can  & 1{,}289 & 46.5\% \\
Tuna fish can    & 1{,}287 & 88.7\% \\
Pudding box      & 1{,}287 & 63.6\% \\
Gelatin box      & 1{,}295 & 83.5\% \\
Potted meat can  & 1{,}238 & 67.8\% \\
Mug              & 1{,}287 & 84.6\% \\
\midrule
\textbf{Pooled}  & \textbf{7{,}683} & \textbf{72.4\%} \\
\bottomrule
\end{tabular}
\end{table}

\subsection{Contact-Anchored Transfer Across Embodiments}
\label{sec:exp_retarget}

The second question is whether contact structure closes the
embodiment gap better than geometric correspondence. We first
isolate the retargeting stage against baselines, then test whether
the advantage survives into closed-loop reconstruction on four
morphologically distinct hands.

We retarget 100~MANO trajectories to DexHand~021 and compare
against DexPilot~\cite{handa2020dexpilot},
dex-retargeting (Position, Vector)~\cite{qin2023anyteleop}, and
Spider~\cite{pan2025spider}.
For Spider, $M$ is the iteration cap of its internal trajectory
optimizer; we report $M\!=\!16$, $M\!=\!32$, and the default
configuration.
We evaluate: \emph{contact fidelity}, the mean finger-to-target
distance; \emph{contact retention}, the fraction of contact frames
where the fingertip reaches the object; \emph{joint-limit
violation rate}; and \emph{joint velocity jitter}
(Table~\ref{tab:retarget}).
Vector achieves the best raw fidelity (10.6\,mm) but
incurs 29.9\% joint-limit violations and 3.46\,rad/s jitter.
Ours is a close second in fidelity (14.7\,mm) with the lowest
jitter (0.53\,rad/s) and only 3.2\% violations---the only method
that jointly achieves high fidelity, high retention, low violations,
and low jitter.

\begin{table}[htbp]
\centering
\caption{Cross-embodiment reconstruction (200~episodes per hand).}
\label{tab:cross}
\vspace{2pt}
\setlength{\tabcolsep}{3pt}
\footnotesize
\begin{tabular}{@{}lcccc@{}}
\toprule
Hand & Replay & Residual & $\Delta$ (pp) & Ratio \\
\midrule
DexHand~021    &  0.5\% & 79.0\% & +78.5 & 158$\times$ \\
DexHand~021Pro &  3.5\% & 80.0\% & +76.5 & 23$\times$ \\
Shadow Hand    & 14.5\% & 82.5\% & +68.0 &  5.7$\times$ \\
Allegro Hand   & 50.0\% & 76.5\% & +26.5 &  1.5$\times$ \\
\midrule
\textbf{Overall} & \textbf{17.1\%} & \textbf{79.5\%} & \textbf{+62.4}
  & \textbf{4.6$\times$} \\
\bottomrule
\end{tabular}
\end{table}

\begin{figure}[htbp]
\centering
\includegraphics[width=0.6\textwidth]{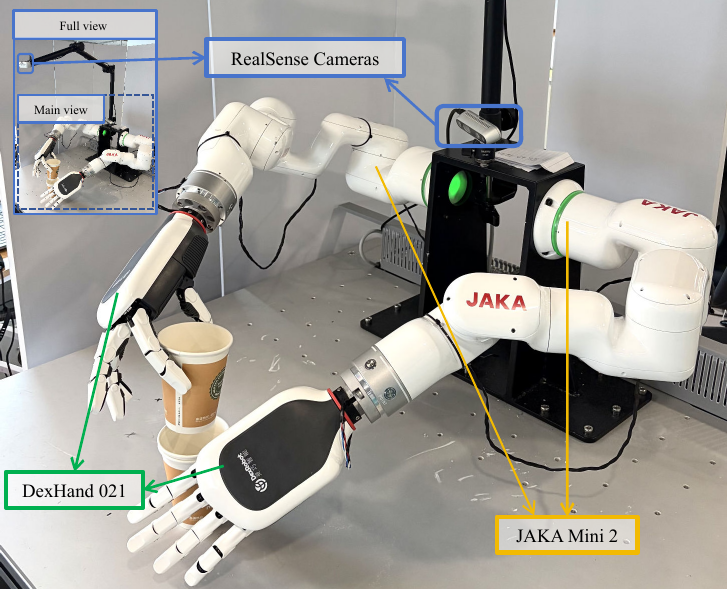}
\caption{Real-world hardware setup: two 6-DoF JAKA Mini~2 arms
  with DexHand~021 and two RealSense D435 cameras.}
\label{fig:real_config}
\end{figure}

Component metrics alone do not imply usable references, so we
next evaluate the retargeted trajectories in closed-loop
reconstruction across embodiments.
\label{sec:exp_cross}

200~single-arm trajectories (3~objects, 7~gestures,
14~object--gesture combinations) are retargeted to four robot
hands---DexHand~021~\cite{yuan2026dexhand}, DexHand~021Pro,
Shadow Hand, and Allegro Hand---with identical training configuration
and no per-hand tuning.
All 200~trajectories are evaluated with and without residual
correction; success requires the object to be grasped, lifted,
and stably placed, with deviations within the same
3\,cm/$30^{\circ}$ threshold.

\begin{figure}[!htbp]
\centering
\includegraphics[width=0.62\linewidth]{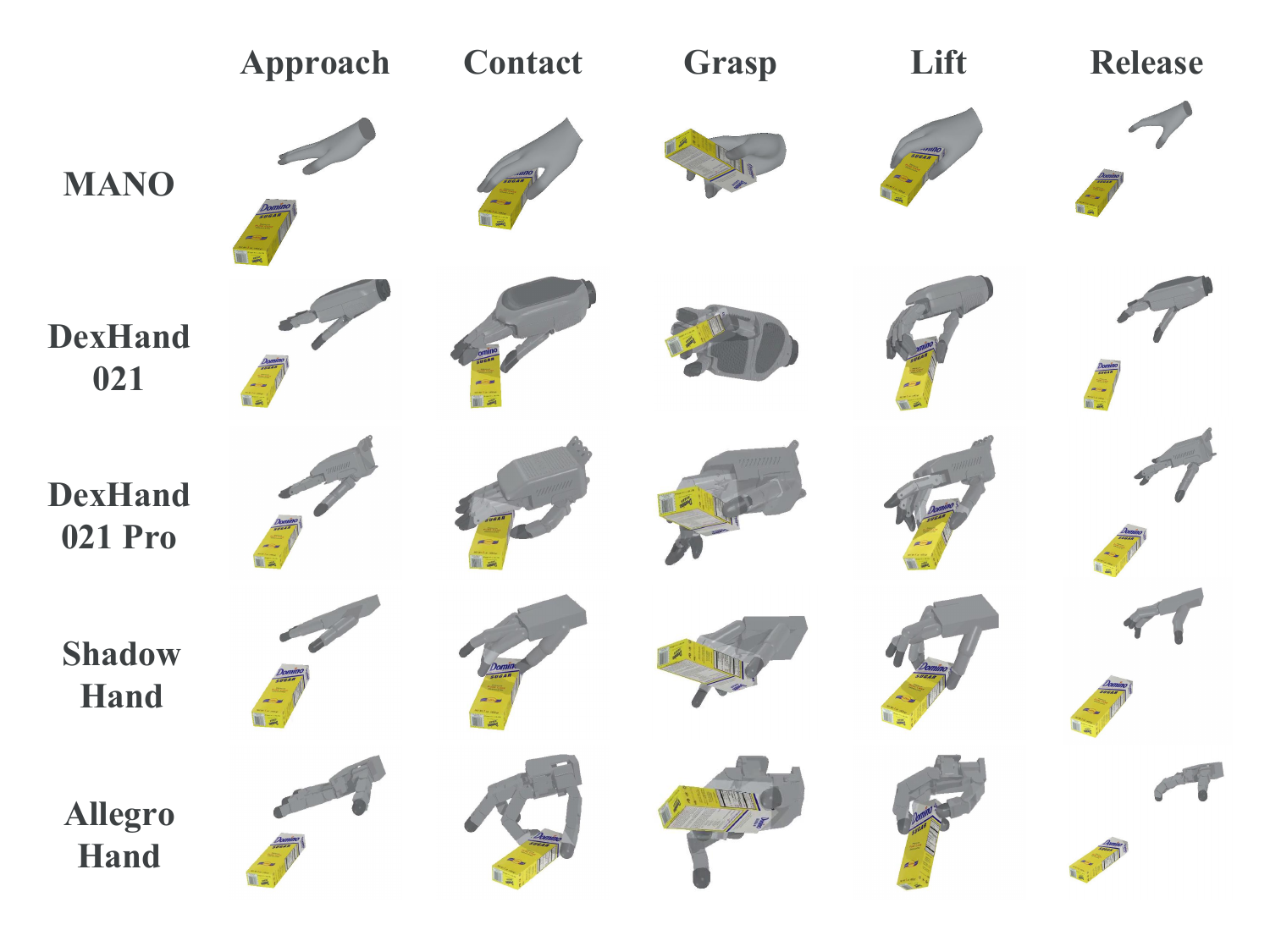}
\caption{\textbf{Cross-embodiment residual policy execution.}
Four morphologically distinct hands (DexHand~021, DexHand~021Pro,
Shadow Hand, and Allegro Hand) execute the same retargeted
grasp-and-lift reference using the residual policy
(Sec.~\ref{sec:residual}). Snapshots show five phases: Start,
Approach, Grasp, Lift, and Release.}
\label{fig:cross_embodiment}
\end{figure}

As shown in Table~\ref{tab:cross}, residual correction raises the
four-hand average from 17.1\% to 79.5\% (+62.4\,pp). Without it,
success varies from 0.5\% (DexHand~021) to 50.0\%
(Allegro Hand); with it, all four hands converge to
76.5--82.5\% (Fig.~\ref{fig:cross_embodiment}), demonstrating
that a single training recipe generalizes across morphologically
distinct embodiments.

\begin{figure}[htbp]
\centering
\includegraphics[width=0.7\linewidth]{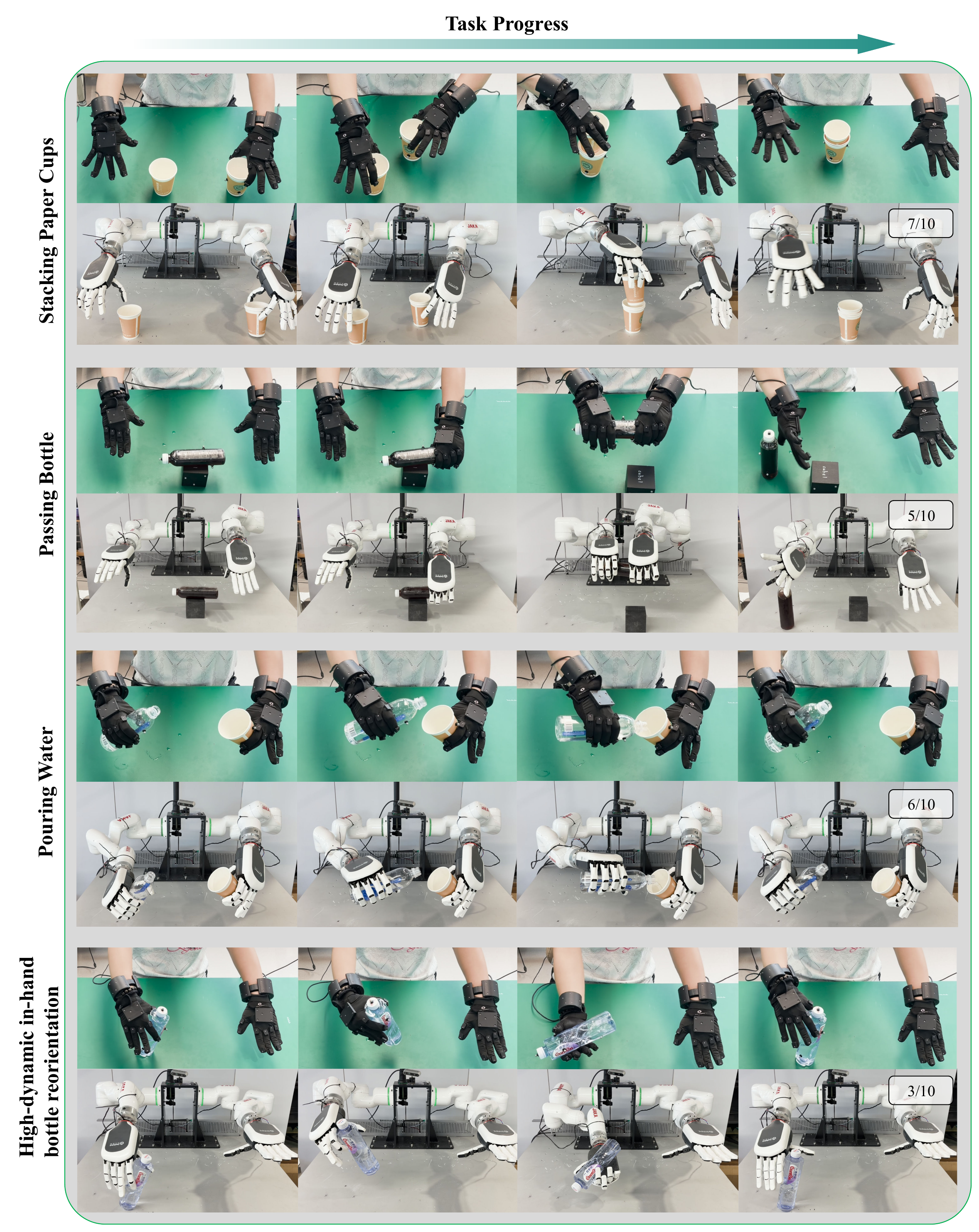}
\caption{Real-world execution of four tasks. Per task, the top row
  shows the human demonstration (green background) and the bottom row
  shows the robot execution, with time progressing left to right.
  From top to bottom: stacking paper cups (7/10), passing \& standing
  a bottle (5/10), pouring water (6/10), and in-hand bottle reorientation (3/10).}
\label{fig:real_exec}
\end{figure}

\subsection{From Data to Deployment}
\label{sec:exp_downstream}

The final question is whether the pipeline's outputs are useful
beyond reconstruction: do recovered contact forces improve
downstream policy learning, and does the complete chain execute on
physical hardware?

We train a Diffusion Policy~\cite{chi2023diffusion} in two
configurations: a 64-dim kinematic baseline and a contact-conditioned
variant that appends 16-dim per-frame contact forces (80-dim total).
Architecture, data, and hyperparameters are shared.
Each of 4~objects $\times$ 4~grasp types (50~trajectories each)
is tested 50~times under a 0.5\,N load and the same
3\,cm/30$^{\circ}$ success criterion (Table~\ref{tab:dp}).
Contact conditioning raises the average from 34.7\% to 46.5\%
(+11.8~pp), with the largest gain on Cube$_2$ (+19.2~pp),
confirming the value of contact information for dexterous
manipulation.

\begin{table}[t]
\centering
\caption{Diffusion Policy with/without contact force
  (0.5\,N load, 3\,cm/30$^{\circ}$ criterion).
  Object subscripts are local to this table and do not correspond to Table~\ref{tab:single}.}
\label{tab:dp}
\vspace{2pt}
\footnotesize
\begin{tabular}{@{}lccc@{}}
\toprule
Object & w/o ct.\ (64-d) & w/ ct.\ (80-d) & $\Delta$ \\
\midrule
Cube$_1$     & 35.2\% & 42.4\% & +7.2 \\
Cylinder$_1$ & 43.6\% & 60.4\% & +16.8 \\
Cube$_2$     & 30.8\% & 50.0\% & +19.2 \\
Cylinder$_2$ & 29.2\% & 33.2\% & +4.0 \\
\midrule
\textbf{Average} & \textbf{34.7\%} & \textbf{46.5\%} & \textbf{+11.8} \\
\bottomrule
\end{tabular}
\end{table}

\begin{table}[t]
\centering
\caption{Retargeting comparison (100 MANO trajectories $\to$
  DexHand~021). Best in bold, second-best underlined.}
\label{tab:retarget}
\vspace{2pt}
\footnotesize
\resizebox{\linewidth}{!}{%
\begin{tabular}{@{}lccccccc@{}}
\toprule
Method & Time (s) & Tput.\ (f/s)
  & Fidelity (mm)$\downarrow$ & Ret.\ $\le$5\,mm$\uparrow$
  & Ret.\ $\le$10\,mm$\uparrow$
  & Jt.\ viol.$\downarrow$ & Jitter (rad/s)$\downarrow$ \\
\midrule
DexPilot~\cite{handa2020dexpilot}
  & 186.8 & 126.2
  & 15.8 & 2.0\% & 35.1\%
  & 21.4\% & 2.94 \\
Position~\cite{qin2023anyteleop}
  & \textbf{21.8} & \textbf{1081.7}
  & 15.7 & 0.3\% & 26.9\%
  & 10.9\% & 1.63 \\
Vector~\cite{qin2023anyteleop}
  & 169.2 & 139.4
  & \textbf{10.6} & \textbf{32.0\%} & \textbf{53.5\%}
  & 29.9\% & 3.46 \\
Spider $M\!=\!16$~\cite{pan2025spider}
  & 3742.0 & 6.3
  & 25.5 & 0.6\% & 5.4\%
  & \textbf{2.66\%} & 3.03\\
Spider $M\!=\!32$~\cite{pan2025spider}
  & 3320.8 & 7.1
  & 25.3 & 0.8\% & 6.8\%
  & \underline{2.68\%} & 3.15 \\
Spider default~\cite{pan2025spider}
  & 1079.3 & 27.5
  & 49.9 & 0.2\% & 1.4\%
  & 3.27\% & \underline{0.73} \\
\textbf{Ours} ($n_{\mathrm{iter}}\!=\!3000$)
  & 8513.2 & 5.5
  & \underline{14.7} & \underline{13.9\%} & \underline{44.6\%}
  & 3.2\% & \textbf{0.53} \\
\bottomrule
\end{tabular}%
}
\end{table}

Finally, we deploy the trained policies on physical hardware.
\label{sec:exp_real}

\textbf{Setup.}
The hardware platform (Fig.~\ref{fig:real_config}) comprises two
6-DoF JAKA Mini~2 arms, each with a DexHand~021, and two Intel
RealSense D435 cameras. A perception stack (YOLOv5, MobileSAM,
FoundationPose~\cite{wen2024foundationpose}) provides real-time
object-pose feedback. Four tasks---stacking paper cups,
passing \& standing a bottle, pouring water, and in-hand bottle
reorientation---are tested over 10~trials each.

\textbf{Success criteria.} Each trial is scored as a binary success
from the recorded RGB video: \emph{stacking paper cups}---the second
cup is placed into the first cup and the stack does not topple;
\emph{passing \& standing a bottle}---the bottle is passed from the
left hand to the right hand, and the right hand then places it on the
table where it remains standing without falling; \emph{pouring
water}---water is poured from the source cup into the target paper
cup without spilling outside it; \emph{in-hand bottle
reorientation}---the bottle is lifted, flipped in hand, and placed
back on the table, where it stands upright without falling.

\textbf{Results.}
As shown in Fig.~\ref{fig:real_exec}, the deployed policies achieve
an overall success rate above 50\%, with stacking cups performing
best (7/10). Bimanual coordination tasks (passing \& standing,
pouring) show stable performance at 5/10 and 6/10 respectively;
in-hand reorientation remains hardest (3/10) due to demanding
continuous re-grasping where small timing errors compound.
These four tasks were selected because they pose particular
challenges for teleoperation systems and parallel-jaw grippers:
each requires multi-finger contact coordination that is difficult
to demonstrate via standard teleoperation interfaces.
All four tasks are executed with frozen
policies and zero real-robot training data, confirming that the
pipeline transfers human demonstrations to physical hardware.


\section{Discussion and Conclusion}

\textbf{Contact as the missing supervision.}
Two independent results point the same way: conditioning a
diffusion policy on recovered contact forces improves success on
every object tested (Sec.~\ref{sec:exp_downstream}), and residual
correction succeeds on exactly the objects where kinematic replay
never does (Sec.~\ref{sec:exp_r2s}), because the policy is trained
to establish and maintain contacts rather than to track joint
trajectories alone. Recovering contact from kinematic recordings is
therefore not a data-enrichment step but a prerequisite for
learning contact-rich manipulation.

\textbf{Why the formulation transfers across hands.}
The retargeting objective depends only on surface geometry, and the
experiments confirm that this suffices: a single human dataset
reconstructs successfully on four morphologically distinct hands
with identical configuration, and the
formulation is the only one among those compared that jointly
achieves high contact fidelity, low joint-limit violations, and low
temporal jitter (Sec.~\ref{sec:exp_retarget}).

We presented a pipeline that converts human motion-capture
recordings into dexterous robot manipulation without real-robot
training data. Physics refinement recovers contact and force
annotations with a single residual-RL policy trained across diverse
tasks; contact-anchored retargeting transfers the demonstrated
contact structure through morphology-agnostic optimization; and
residual policy learning restores dynamic feasibility for direct
hardware transfer. Because every stage after capture runs in
simulation, the pipeline scales with the availability of human
demonstrations rather than robot access.

\textbf{Toward in-context policy learning.}
The one-to-many structure of the generated data, in which one human
demonstration is paired with many robot trajectories under varied
initial conditions, naturally provides in-context examples for
policies that map a human video prompt to condition-dependent robot
actions. By perturbing initial object and hand poses in simulation,
each demonstration yields diverse robot executions that share the
same task intent but differ in starting configuration, aligning
directly with emerging in-context learning frameworks where a model
conditions on a demonstration prompt and acts on the current state.
Connecting this data engine to in-context policy architectures is
therefore a natural next step toward robots that acquire new
dexterous skills from a single human demonstration.

\bibliographystyle{abbrvnat}
\bibliography{refs}

\end{document}